\documentclass[letterpaper, 10 pt, conference]{ieeeconf}  
\usepackage{etoolbox}
\makeatletter
\patchcmd{\@makecaption}
  {\scshape}
  {}
  {}
  {}
\makeatother

\IEEEoverridecommandlockouts
\usepackage{microtype}     
\usepackage{hyperref}
\usepackage{url}
\usepackage{mwe}
\usepackage{float}
\usepackage{multirow}
\usepackage{indentfirst}
\usepackage{amssymb}
\usepackage{amsmath}
\usepackage{tabularx}
\usepackage{array}
\usepackage{diagbox}
\usepackage{array}
\usepackage{booktabs}
\usepackage[vlined,ruled,commentsnumbered]{algorithm2e}
\usepackage{algpseudocode}
\usepackage{amsmath}
\usepackage{graphicx}
\usepackage{subfigure}
\usepackage{epsfig}
\usepackage{verbatim}
\usepackage{wrapfig}
\usepackage{textcomp}
\usepackage{graphics} 
\usepackage{epsfig} 
\usepackage{mathptmx} 
\usepackage{times} 
\usepackage{float}
\usepackage{cite}
\usepackage[OT1]{fontenc} 
\title{\LARGE \bf
Transferable Force-Torque Dynamics Model for Peg-in-hole Task
}

\author{Junfeng Ding$^{*}$, Chen Wang$^{*}$, Cewu Lu
\thanks{$^{*}$ These authors have contributed equally.}
\thanks{All authors are with the Department of Computer Science, Shanghai Jiao Tong University. {\tt\small \href{mailto:newuhe@sjtu.edu.cn}{newuhe},\href{mailto:jere.wang@sjtu.edu.cn}{jere.wang},\href{mailto:lucewu@sjtu.edu.cn}{lucewu}@sjtu.edu.cn}. Junfeng Ding and Cewu Lu are also at Flexiv Robotics Research.}
}

\begin{document}

\maketitle
\thispagestyle{empty}
\pagestyle{empty}

\begin{abstract}

We present a learning-based force-torque dynamics to achieve model-based control for contact-rich peg-in-hole task using force-only inputs. Learning the force-torque dynamics is challenging because of the ambiguity of the low-dimensional 6-d force signal and the requirement of excessive training data. To tackle these problems, we propose a multi-pose force-torque state representation, based on which a dynamics model is learned with the data generated in a sample-efficient offline fashion. In addition, by training the dynamics model with peg-and-holes of various shapes, scales, and elasticities, the model could quickly transfer to new peg-and-holes after a small number of trials. Extensive experiments show that our dynamics model could adapt to unseen peg-and-holes with 70\% fewer samples required compared to learning from scratch. Along with the learned dynamics, model predictive control and model-based reinforcement learning policies achieve over 80\% insertion success rate. Our video is available at \url{https://youtu.be/ZAqldpVZgm4}.
\end{abstract}

\section{Introduction}
Insertion operation is one of the basic robotic manipulation skills which has a wide range of applications in industrial assembling and housework. Recent studies use rule-based strategies based on contact state analysis \cite{lefebvre2005online,jakovljevic2012contact,abdullah2015force,yu2018realtime} or model-free reinforcement learning \cite{levine2016end,8202244,lee2018making} to solve the task. However, these methods require either human knowledge or expensive online learning on real robots. To achieve sample-efficient policy learning, model-based control methods are introduced. \cite{levine2015learning,fu2016one} train the dynamics of proprioception, such as joint configuration and velocity, followed by model-based RL algorithms. Nevertheless, these learned dynamics do not incorporate contact-rich force information, which hinders their performance in high-precision insertion challenges. In this work, we study the dynamics model based on the end-effector force-torque feedback, a contact-rich information source for solving peg-in-hole task.

\begin{figure}
\centering
\label{fig:robot_setting}
\includegraphics[width=1\linewidth]{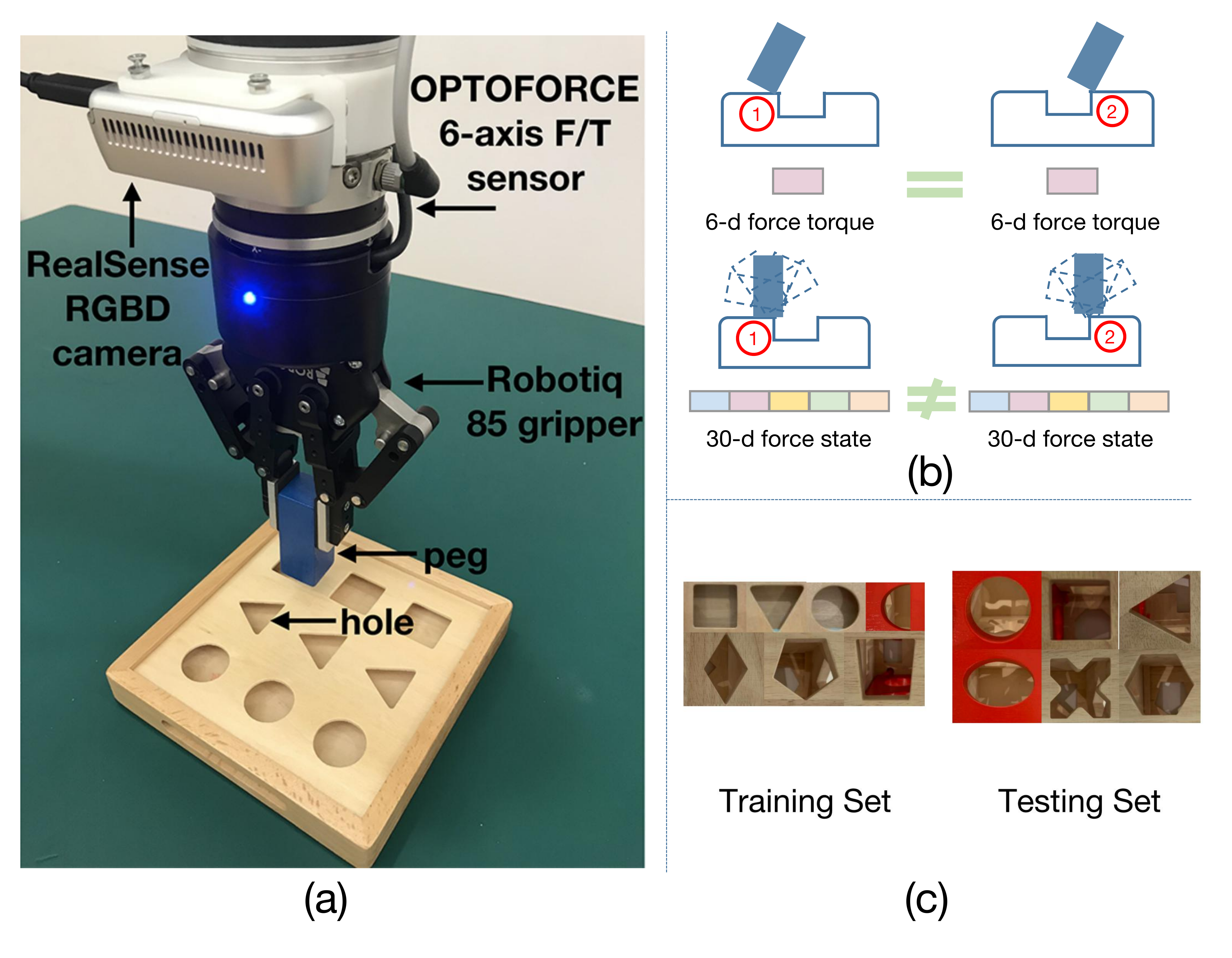}
\caption{(a) Illustration of robot hardware. The RGB-D camera is used for initial rough alignment, and the inserting process only uses force-torque feedback. (b) Single 6-d force-torque signal is ambiguous, while the 30-d force state acquired by the multi-pose sampling is distinct. (c) Examples of the pegs and holes used for training and testing.}
\label{fig:setting}
\end{figure}

However, modeling the force-torque dynamics is challenging due to three reasons: 1) Single 6-dimensional force-torque signal is ambiguous, that is: force-torques at different contact positions are too similar to distinguish the contact locations (Fig. \ref{fig:setting}(b)). 2) Data-driven dynamics learning requires an extensive dataset, which is expensive and time-consuming in real robot scenarios. 3) The learned dynamics model should have the ability to transfer to new pegs and holes with unseen characteristics, such as changes in shape, scale, and elasticity.

We focus on tackling these problems and propose a transferable force-torque dynamics model for peg-in-hole task. To deal with the ambiguity of the force state, we enrich the force-torque feedback by multi-pose sampling at each contact position and regard the concatenation of the feedbacks as one force state. This is inspired by human insertion, where the human operator tends to rotate the peg as it touches the edge of the hole to acquire more information for localizing the peg. In this way, the enriched force representation of different contact locations could then be distinguished easily (Fig. \ref{fig:force_pattern}(a)). Based on this state representation, a learning-based dynamics model is trained to predict the next state given the current force state and action. Instead of collecting a large number of training data as \cite{levine2016end,8202244} for the dynamic learning, an offline data generation method is proposed, which generates numerous synthetic trajectories based on a small set of grid-sampled force-torque feedbacks (Fig. \ref{fig:dynamics_training}). These trajectories are then fed to the dynamics model for supervised training. In order to achieve fast adaptation to unseen pegs and holes, the training set of the dynamics is a mixture of pegs and holes with a variety of characteristics, i.e. difference in shapes, scales, and elasticities (Fig. \ref{fig:setting}(c)). Finally, with the learned dynamics model, model predictive control \cite{garcia1989model} or model-based RL is applied to complete the insertion.

Our algorithm is proven capable of completing peg-in-hole task sample-efficiently on a UR5 robot with an OPTOFORCE force-torque sensor.
The primary contributions of this work are three-fold:
\begin{enumerate}
\item A force-torque dynamics model learned on a
multi-pose force-torque state representation, with which two model-based control schemes are performed to complete the peg-in-hole task successfully.
\item An offline data generation method that synthesis numerous data from a small set of samples, ensuring the force-torque dynamics could be learned sample-efficiently.
\item A training set collected with pegs and holes of various characteristics. The dynamics model trained on this dataset has efficient transferability to unseen cases.
\end{enumerate}

\vspace{1\baselineskip}
\section{Related Work}
This paper focuses on solving the problem of peg-in-hole assembly using learning-based force-torque dynamics model. Thus, this section reviews the related work in peg-in-hole problem and the literature of dynamics learning.

\subsection{Peg-in-hole challenge}

The peg-in-hole insertion is a problem that has been extensively examined in the scientific literature. Traditional methods start from a theoretical point of view \cite{li1997hybrid, qiao1993robotic, qiao1995fine}, where researchers made efforts to analyze the contact model between the peg and hole and developed strategies based on that. To improve the insertion accuracy, most conventional methods tend to optimize the contact state estimation through analytical methods \cite{fei2003assembly,lefebvre2005online,xia2006dynamic} and statistical techniques \cite{lefebvre2005online,gadeyne2005bayesian,jakovljevic2012contact,abdullah2015force,yu2018realtime}. After acquiring the contact state, compliant control strategies \cite{lefebvre2005active,brignone2002geometrically,tang2016autonomous,hou2018learning,977187} are applied to combine the mating parts. Since the contact states are predefined for specific kinds of holes, these methods fail to accommodate to pegs and holes with variant characteristics in the real world.

Recent learning-based methods tend to remove the specific engineering of the feedback controller. Supervised learning methods train the policy model to imitate human strategies by learning from demonstrations (LfD) \cite{wan2017optimal, tang2015learning, tang2016teach}. To avoid the bias of human policies, model-free reinforcement learning algorithms are used to handle insertion problems based on feedback from proprioception \cite{levine2016end}, force \cite{8202244, kalakrishnan2011learning, sung2017learning, xu2018feedback} and the combination of haptic and vision \cite{lee2018making, van2016stable}. Although these methods show promising results for this task, a sample-inefficient online learning procedure makes training on real robots expensive and time consuming. Model-based RL methods \cite{levine2014learning,levine2015learning,fu2016one} reduce the required interaction time by offline training the system dynamics. However, these methods take proprioception as a state representation, ignoring the contact-rich information from force-torque feedback. We propose a learning-based force-torque dynamic model for tackling the peg-in-hole challenge.

\subsection{Dynamics learning}
Model-based algorithms must be equipped with a model that can represent a good approximation of the dynamics to reduce the required interaction time. \cite{8794219} learned a tactile predictive model to conduct model predictive control \cite{garcia1989model} for manipulation. Nevertheless, sufficient experience must be provided to optimize the model to produce accurate dynamics predictions \cite{polydoros2017survey}. This challenge can be mitigated by incorporating domain knowledge about the system \cite{nguyen2010using,cutler2015efficient}. \cite{wu2010towards} used demonstrations, and \cite{deisenroth2011pilco} relied on the assumption of smoothness to learn the dynamics. 
A guided policy search was proposed \cite{levine2014learning,levine2015learning,montgomery2016guided} using time-varying linear-Gaussian to fit local dynamics iteratively, with a background dynamics distribution acting as a prior to reduce the sample complexity. However, these models all need to sample online for local dynamics fitting. \cite{fu2016one} learned a coarse global dynamics model and locally adapted it online to unseen tasks. However, it only used proprioception features and required too much training data for the global dynamics model. To facilitate dynamics learning, we use force-torque information and construct a learnable force state.

\section{Problem Setting}

The core of our work is a force-torque dynamics model that can be used by model predictive control and model-based RL to perform the peg-in-hole task. Additionally, the learned dynamics model should have the ability to rapidly adapt to new pegs and holes with unseen characteristics, such as variations in shape, scale and elasticity. By considering the peg-in-hole task as finite-horizon episodic, the dynamics model learns to predict the state transformation $x_{t+1} = D(x_t, u_t)$, where $x_t \in X$ is the state at time step $t$ within the state space $X$ and $u_t \in A$ is the action at time step $t$ given the action space $A$. This dynamics model is presented by a neural network with parameters $\theta$ that are trained on a large collected force-torque dataset that contains pegs of seven common shapes with different characteristics as described in Sec. \ref{sec:fast}.

The control system aims to minimize the total cost $\Sigma_{t=1}^{T}{L(x_t, u_t)}$, where $T$ is the time horizon and $L$ refers to the cost function. Based on the learned dynamics model, we use the model predictive control (MPC) algorithm \cite{garcia1989model} to optimize the planned action trajectories with respect to the cost function, which is introduced in Sec. \ref{sec:mpc}. In addition, the dynamics model can also be used for off-line training a model-based reinforcement learning policy model $\pi: X \rightarrow P(A)$. The model-based RL is detailed in Sec. \ref{sec:model-based}.

In this work, the insertion pipeline is divided into a two-step process: (1) vision-based rough align, and (2) torque-based precise align. Rough alignment can be simply realized by moving the robot arm TCP position to the nearest deepest hole on the depth image provided by the end-effector RGB-D camera. This work mainly focuses on the second torque-based precise insertion process. The clearance of all the training and testing peg-hole pairs is 1 ($\pm$ 0.5) mm. We empirically find that vision-based rough alignment can ensure the initial align error within the range of 2 ($\pm$ 0.5) mm, where the peg is close enough to the hole so that the end-effector torque feedback is rich enough to complete the insertion.

\begin{figure*}[ht]
    \centering
        \includegraphics[width=1\textwidth]{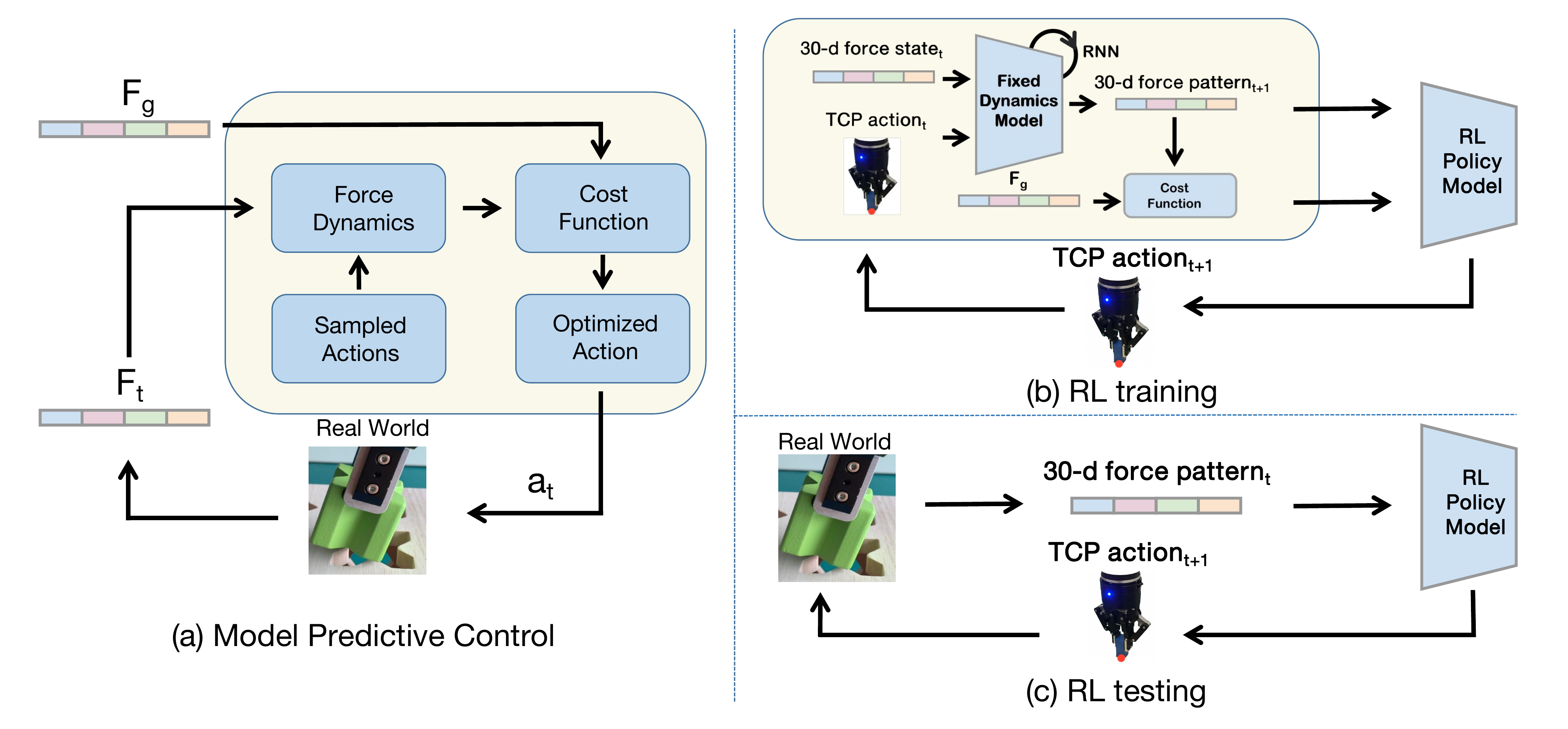}
        
    \caption{(a) The workflow of model predictive control. Given the current force state and a learned dynamics model, we can predict the outcomes for the sampled action sequence. At each time step, the algorithm samples multiple actions and computes their cost, which is computed based on the difference between the predicted force state and the goal state. The first action of the action sequence with the lowest cost is executed. (b) The training process of model-based reinforcement learning. The RL policy obtains the force state from the dynamics model and outputs a discrete action. They are fed to the dynamics model, and the next force state is outputted. The reward is computed between the current force state and the goal state.
    (c) The testing process of model-based reinforcement learning. The robot conducts multipose insertion sampling and obtains a force state. The RL policy takes the force state as input and outputs a discrete action. Then, the robot takes action and samples again to obtain the next force state for the next action.}
    \label{fig:overflow}
\end{figure*}

\begin{figure}
    \centering
    
        \label{fig:robot_setting}
        \includegraphics[width=1\linewidth]{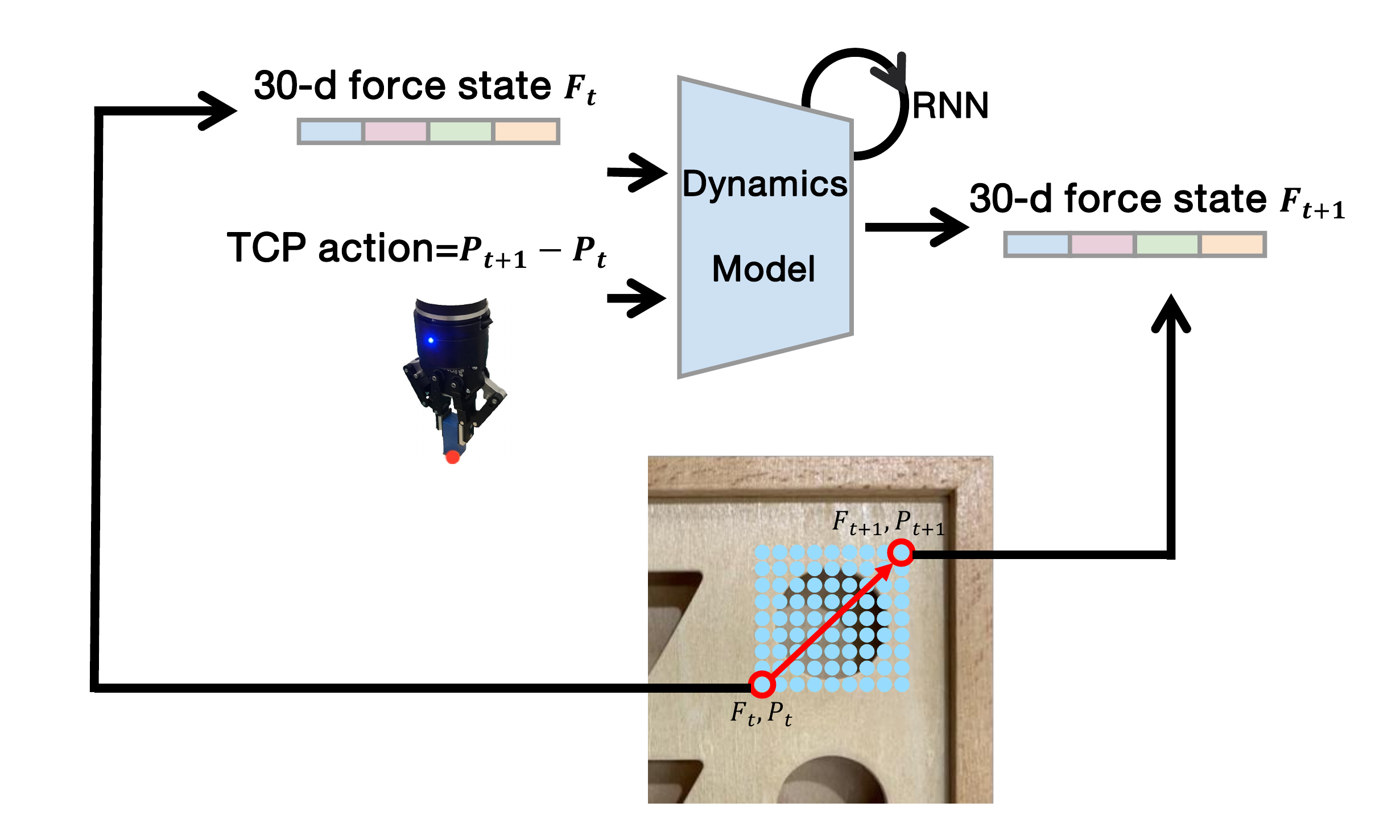}

\caption{Training process of the dynamics model. Training data are randomly generated from the force-torque value and TCP information of the grid-sampled positions on the hole. The dynamics take a 30-d force state and current action as input and output the force state of the next step. The training is supervised.}
\label{fig:dynamics_training}
\end{figure}

\begin{figure}[h]
\begin{center}

\subfigure[Force pattern at different positions of the same hole ]{
\includegraphics[width=0.45\textwidth]{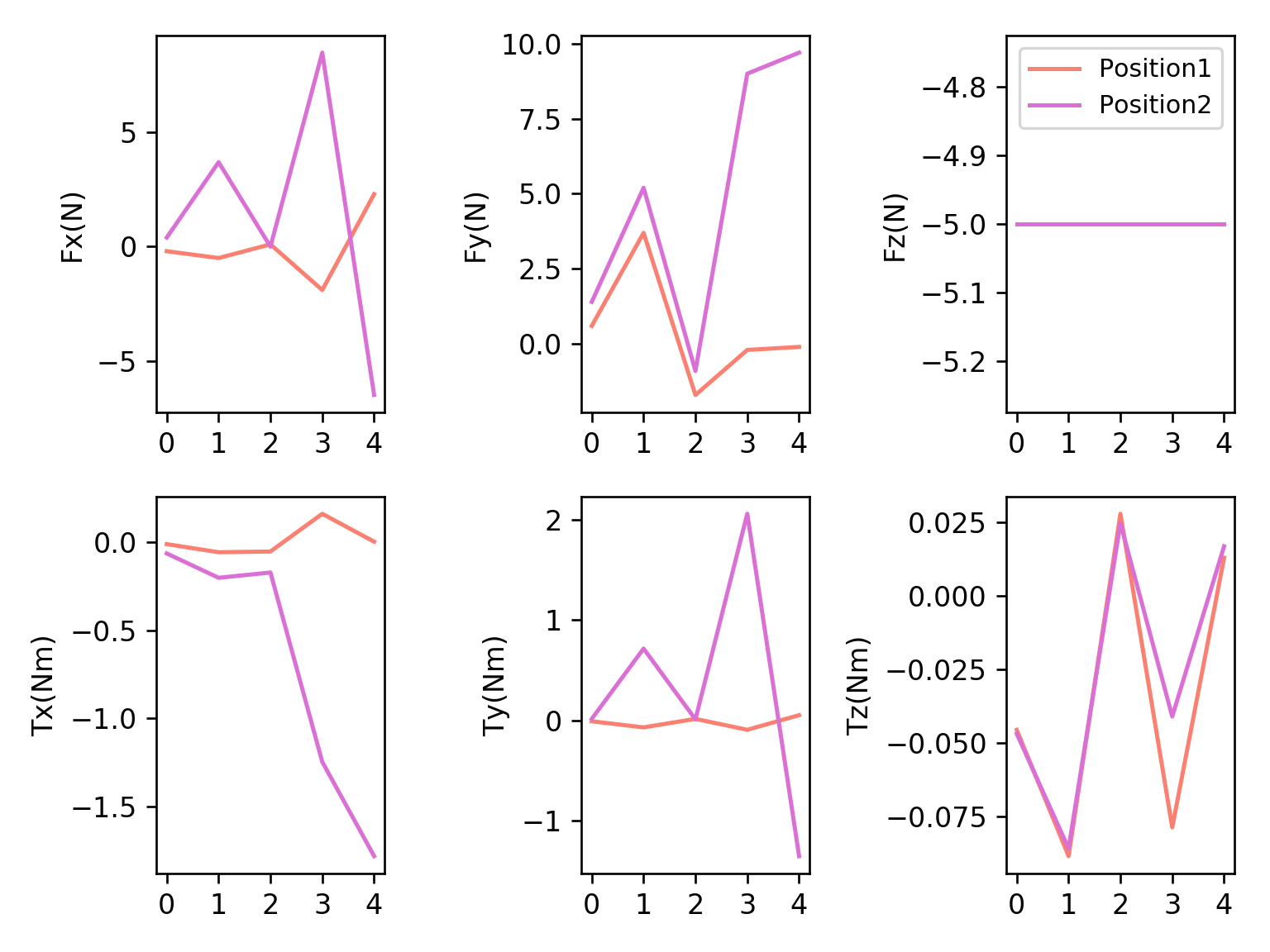}
}
\subfigure[Force pattern at the same position of different holes]{
    \includegraphics[width=0.45\textwidth]{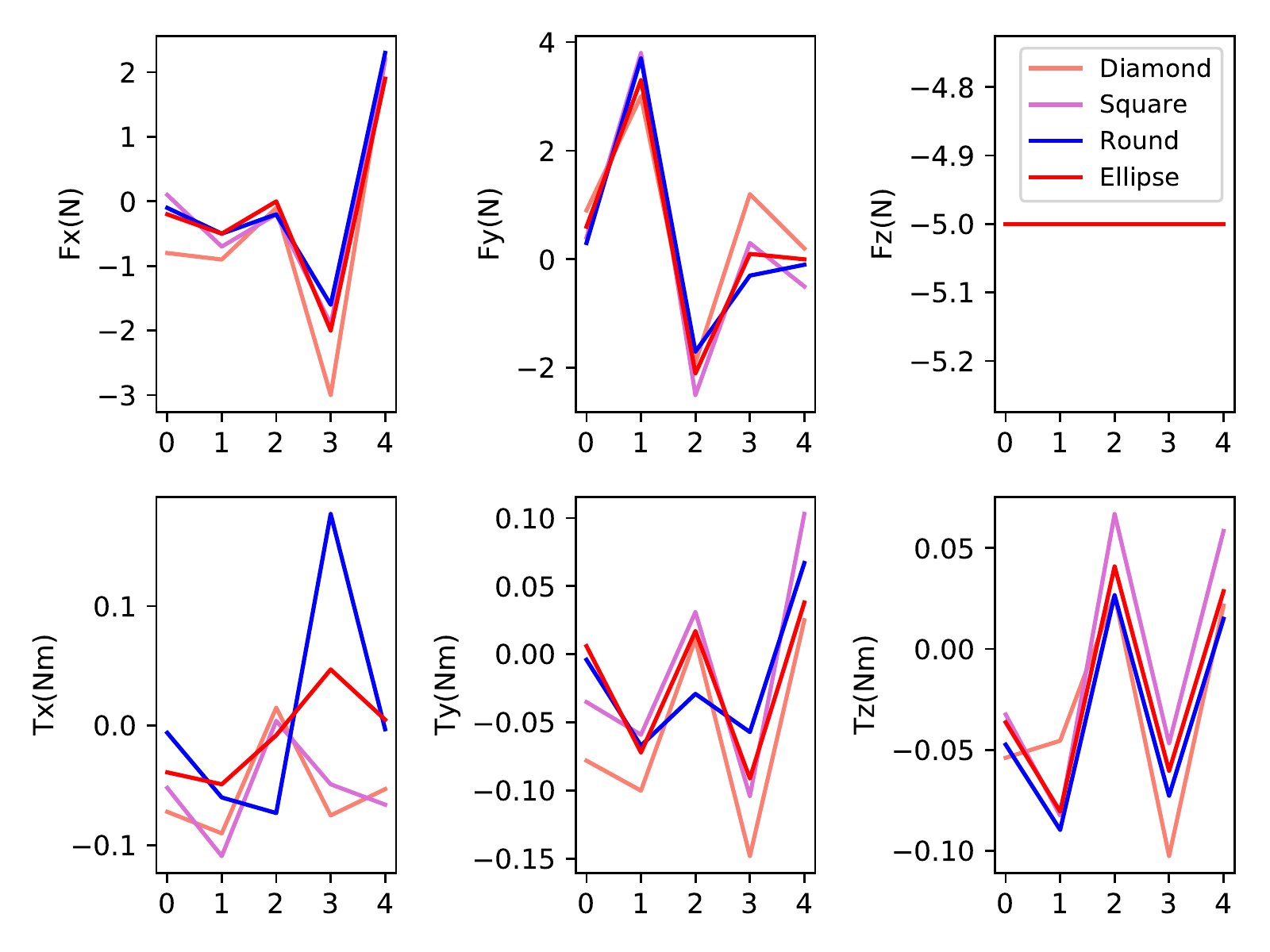}
}
\end{center}
\caption{The graph shows the multipose force-torque state representation. (a) The force pattern at different contact positions of the same round hole is different. (b) The force representation at the same contact position of four different holes shares a similar pattern. } 
\label{fig:force_pattern}
\end{figure}

\section{Modeling}
Our model is composed of five parts: the multi-pose force-torque state representation, the dynamics model learning process using offline data generation, transferability to unseen pegs and holes and two model-based control methods: model predictive control and model-based reinforcement learning.

\subsection{Multi-pose Force-Torque State Representation}
The end-effector 6D force-torque signal consists of both the force and torque along the 3-axis (x-y-z). Single force-torque feedback is ambiguous for precisely indicating the location of the peg relative to the hole. To overcome this limitation, we propose a multi-pose force-torque state representation to increase the information gain. For each position, the robot holds the peg in five different end-effector poses and continues moving straight down until the z-axis force meets the stop threshold $\emph{f}_{max}$. Then, the TCP positions and corresponding torque force feedback $f_t$ are recorded. After that, the collected five single 6-d force-torque vectors are concatenated into a 30-dimensional vector which is regarded as the force-torque state representation $F_t$ in our method. Here the first 15 dimensions of $F_t$ are the force measurement $F^{force}$ and the last 15 dimensions are the torque $F^{torque}$. The five end-effector poses consist of one pose with 0-degree rotation and four poses with $\pm$ $\delta$-degree rotation in the x-axis and y-axis. In our experiment, we select $\delta = 30$ to achieve the best performance and efficiency. By concatenating low dimensional 6D force signals into one 30-dimensional feature vector, we successfully enrich the state representation so that the force-torque state at different relative positions can be easily distinguished (Fig. \ref{fig:force_pattern}(a)). Thus, the dynamics of the force state is learnable.

Considering the transferability to unseen peg-hole pairs is vital for the real-world practice of the dynamics model, we also want to find out whether the enriched force-torque representation has the potential to transfer between peg-and-hole pairs with different characteristics. Ideally, when the pegs with different scales and shapes contact the same relative position to the hole with the same end-effector pose, the torque state representations should share some common feature patterns. We try four different peg-hole pairs and visualized the constructed multi-pose force state in Fig \ref{fig:force_pattern}. Six subfigures separately display the 6-d force-torque reading: Fx, Fy, Fz, Tx, Ty, Tz and the horizontal axis represents five sampling poses. Four curves with different colors show the force state of four different pegs at the same relative position. It is obvious that the multi-pose force state is completely different at different positions, and they share similar patterns between different holes. This is a positive factor not only for learnability but also for the transferability of our learning-based torque dynamics model.

\subsection{Dynamics Learning using Offline Data Generation}
\label{sec:dynamic}
To reduce the number of real-world interactions, we apply an offline training method which generates numerous trajectories from the collected samples. The whole training process of the dynamics is summarized in Fig \ref{fig:dynamics_training}.
First, a discrete sample collection procedure is performed. An N*N grid searching space ($R_x$ in the x-axis and $R_y$ in the y-axis) is first formed around the holes and the 30-d torque state of each point position is recorded in the dataset. 
Second, a random action sequence $(a_1,a_2,\cdots,a_{T})$ is generated starting from an initial position $s_1$ randomly selected from the sampled grid on a hole. At each time step $t$, $a_t$ is composed of a gripper position change in the x- and y-directions, denoted $a_x$ and $a_y$, which are sampled from a Gaussian distribution with zero as the mean and $L_x/2,L_y/2$ as the variance. Third, given the initial position and the action sequence, the trajectory is acquired. For each position on the trajectory, its corresponding ground truth force-torque state is the 30-d force-torque vector of the nearest sampled grid point. In this way, the ground truth force torque state sequence of the generated trajectory $(F_1,F_2,\cdots,F_{T})$ is also achieved. At last, the dynamics model consisting of a 2-layer long short-term memory network \cite{hochreiter1997long} is then supervised trained with the collected data. It takes the force state sequence $(F_1,F_2,\cdots,F_{T})$ along with the action sequence $(a_1,a_2,\cdots,a_{T})$ as input and outputs the state prediction sequence $(\hat{F_2},\hat{F_2},\cdots,\hat{F}_{T+1})$. 
The loss function of the dynamics network $D_{\theta}$ is given by
\begin{equation}
{L}=\frac{1}{T}\sum_t \left \| (F_{t}- \hat{F}_{t+1}) \right \|^2
\end{equation}

\subsection{Transferability to Unseen Peg-and-holes}
\label{sec:fast}
To realize the fast adaptation to unseen pegs and holes, pegs and holes in the training dataset are varied in two aspects to cover different characteristics. First, different shapes of the peg-hole pairs are selected, which include square, round, triangle, ellipse, diamond, trapezium and pentagon. Second, the scale is varied. For example, the diameters of the round holes used are 10 mm, 20 mm and 30 mm. The edge lengths of the square holes are of 10 mm, 20 mm and 30 mm. Note that the selected holes in the training set are all made up of deformable thin boards. For the test data, we choose holes with different shapes, sizes and elasticities to verify the transferability of the learned dynamics model. For instance, there is an ellipse hole with a long radius of 15 mm and an X-shaped hole and they are all built with rigid material. The dynamics is initialized using the training process in Sec \ref{sec:dynamic} with these pegs and holes. When encountered unseen cases, it is re-trained using trajectories produced by the offline data generation method, which only requires a small number of trials. Then the dynamics has the ability to predict the force state in this new case.

\subsection{Model Predictive Control}
\label{sec:mpc}

Once the learned dynamics model is acquired, the model predictive control (MPC) \cite{garcia1989model} can be used to perform the peg-in-hole task. The planning process is illustrated in Fig. \ref{fig:overflow}(a). The goal state $F_g$ is specified as the 30-d torque state of the inserted position. A random action sequence $(a_1,a_2,\cdots,a_{T})$ is then generated. The 2-d continuous action $a_t$ is composed of gripper position change $a_x$ in the x-direction and $a_y$ in the y-direction. They are sampled from a Gaussian distribution with zero as the mean and $R_x/2,R_y/2$ as the variance. Given the action sequence, the force dynamics are utilized to predict the force prediction sequence.
To choose the best action sequence, the difference between its force prediction and the goal force state is computed. The cost function measuring the difference between the predicted single force state $\hat{F_t}$ and the goal $F_g$ is given by
\begin{align*}
    C_t(\hat{F_t},F_g)= \alpha \left \| F_g^{force}- \widehat{F}_t^{force} \right \|^2 + \beta \left \| F_g^{torque}- \widehat{F}_t^{torque} \right \|^2
\end{align*} 
Here $\alpha$ and $\beta$ make the force and torque value in the force state on the same scale. In our experiment, $\alpha$ is empirically set at 0.05 and $\beta$ is set at 1.
Then the optimization criterion for the force state sequence is given by
\begin{align*}
    a_{1:T}=\arg \min_{a_{1:T}} \sum_{t=1,\cdots,T} C_t(\hat{F_t},F_g)
\end{align*}
At each step t, the first action of the chosen best action sequence is executed. Specifically, the stochastic optimizer is cross-entropy \cite{de2005tutorial}. The sample number is set to 200 and the planning horizon $T$ is set to 6.

\subsection{Model-based Reinforcement Learning}
\label{sec:model-based}

With the learned dynamics model, the training of the RL policy model can be changed from online to offline. That is, the transition information and the reward can be fetched from the fixed dynamics model rather than the real-world environment. The training process is illustrated in the Fig. \ref{fig:overflow}(b). At each time step $t$, the policy network made up of a 2-layer MLP takes the 30-d force-torque representation $F_t$ from the dynamics model as input and outputs a discrete end-effector action $a_t$. The action space of the RL policy model consists of one unit movement in eight directions: east, south, west, north, northeast, southwest, southeast and northwest. Then $a_t$ is decomposed into gripper position change $a_x,a_y$ in the x- and y-directions, as mentioned previously. $F_t$ and $(a_x,a_y)$ are fed into the fixed dynamics model for the force-torque state $F_{t+1}$ of the next timestep. 
The cost function is computed using the current state $F_t$ and the goal force state $F_g$:

\begin{equation}
L=\left\{
\begin{array}{rcl}
1       &      & {G(F_t,F_g)>\epsilon}\\
-0.02     &      & {G(F_t,F_g)<\epsilon}
\end{array} \right. 
\end{equation}
where $G(F_t,F_g)=exp(-(F_t-F_g)^2/\sigma)$, and $\epsilon$ is 0.9 in our experiment. $F_g$ is specified as the 30-d force state of the inserted position, the positive reward is the goal reward and the negative reward allows the agent to finish the task as soon as possible.

The RL testing process is shown in the lower right of Fig. \ref{fig:overflow}(c). The learned RL policy takes the current force state from the real world and outputs a discrete action $a_t$ in the same form as that in the training process. The robot holding the peg changes its end-effector position following $a_t$ and conducts multi-pose insert sampling to obtain the next force state.

\begin{figure*}[ht]
    \centering
        \includegraphics[width=1\textwidth]{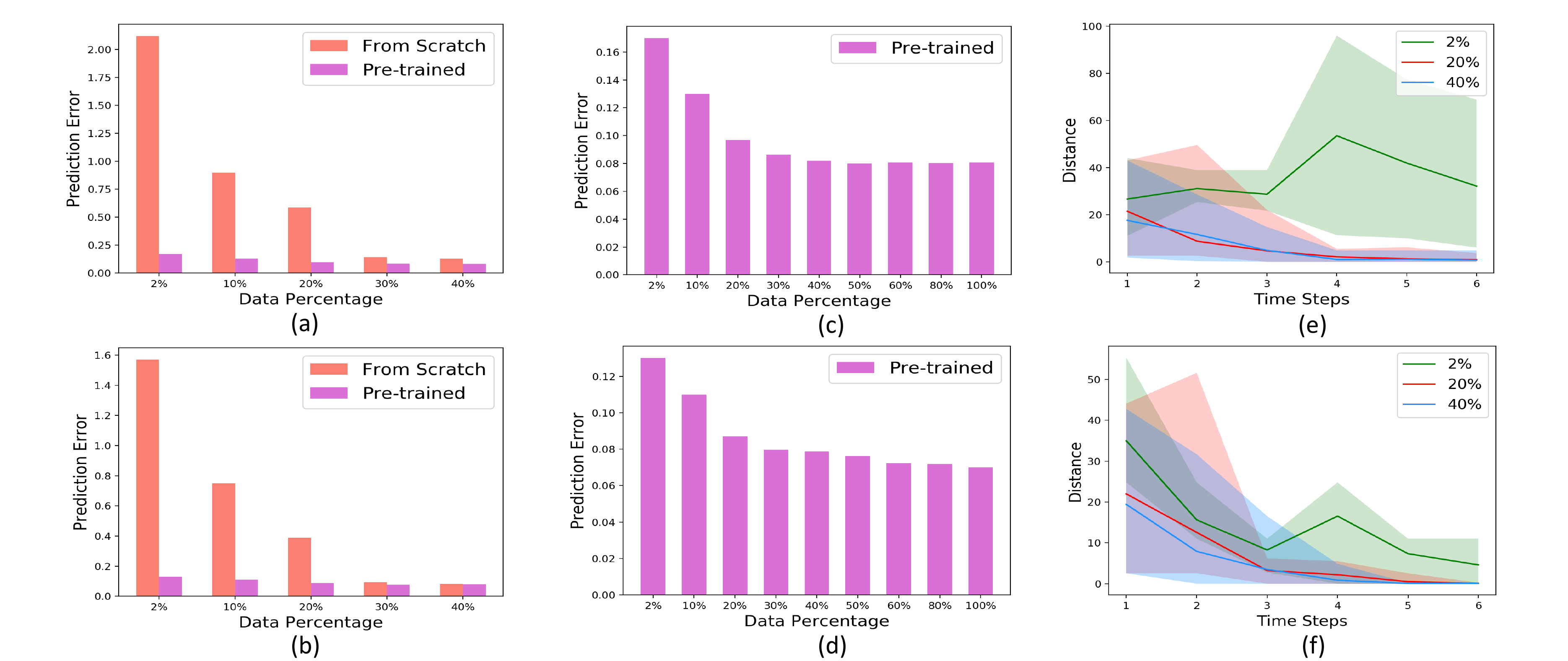}
    \caption{
    (a) shows the force prediction error of the dynamics model from scratch, and the model pretrained and finetuned using 2\%, 20 \% and 40\% of all 81 points data collected in the unseen ellipse hole and (b) shows that finetuned in the new round hole.
    (c) shows the force prediction error
    of the pretrained model finetuned using from 2\% to 100\% of all 81 point data collected in the unseen ellipse hole and (d) shows that finetuned in the new round hole.
    (e) and (f) show the performance of the MPC and model-based RL policy using the dynamics models. The green, red, and blue lines represent the performance of policies trained with the dynamics models finetuned using 2\%, 20 \% and 40\% of all 81 points of data, respectively.
    }
    \label{fig:dynamics_result}
\end{figure*}

\section{EXPERIMENTS}
\label{sec:result}
\subsection{Setup} 

\textbf{Pegs and Holes.} The training dataset for the dynamics model is composed of pegs and holes of different shapes: square, round, semicircle, triangle, diamond, pentagon, trapezium pegs and holes (Fig. \ref{fig:setting})(c). Each shape has 3 different sizes. The edge lengths of the square, triangle, diamond, trapezium and pentagon hole are 10, 20 and 30 mm. The diameters of the round and semi-circle hole are 10, 20 and 30 mm. They are all made up of deformable thin boards.
The testing set includes holes of unseen shapes: ellipse, hexagon, L-shaped and X-shaped. The edge length of hexagon, L-shaped and X-shaped holes is 15 mm. There are also round, square and triangle holes of unseen size (15 mm). They are all made up of rigid material and are hard to deform by contact. All peg-and-holes have a clearance of approximately 1 mm.

\textbf{Robot.} We use UR5, with Robotiq 85, a two-finger gripper with an 8.5cm stroke to hold peg (Fig. \ref{fig:setting}(a)). The robot arm is under TCP position control and controlled using the URScript API via a TCP socket, which embeds the inverse kinematics algorithm.

\textbf{Force Sensor.} Our force-torque sensor is OPTOFORCE 6-axis F/T sensor HEX-E attached to the end-effector of the robot's gripper (Fig. \ref{fig:setting}(a)), which also communicates with the computer via a TCP socket. It has a force resolution of 1/160 N and torque resolution of 1/2000 N·m. The values are biased to zero the force and torque reading. Its sample rate is set at 500 Hz, and the signal filtering cut-off frequency is set at 15 Hz.  

\textbf{Vision Sensor.} An Intel RealSense D435 depth camera is mounted on the robot’s wrist (Fig. \ref{fig:setting}(a)) for the initial rough alignment.

\begin{table}
\begin{center}
\caption{Success rate for inserting in the test holes using MPC or model-based reinforcement learning.}
\begin{tabular}{ |  p{1.5cm}<{\centering} | p{1cm}<{\centering}  | p{2cm}<{\centering}  | p{1cm}<{\centering}  | p{1cm}<{\centering}  | }
  \hline
  Task & MPC & model-based RL \\ \hline
  round & 0.93 & 1  \\ \hline
  square & 0.87 & 0.98  \\ \hline
  triangle & 0.81 & 0.95  \\ \hline
  ellipse & 0.85 & 0.97  \\ \hline
  hexagon & 0.83 & 0.92  \\ \hline
  X-shaped & 0.80 & 0.89 \\ \hline

\end{tabular}
\label{tab2}
\end{center}
\end{table}
    
\subsection{Dynamics Prediction}
First, we evaluate the prediction accuracy of our dynamics model.
Our dynamics model is pretrained using all 7 holes in the training set and each hole is sampled 9*9=81 points to generate 10-step trajectories. Then the dynamics model is trained using generated trajectories. The training process takes 16k episodes and each episode uses 20 trajectories. Then it is finetuned using trajectories generated from 16 points sampled on the X-shaped and round hole in the test set. The finetuning process for the X-shaped hole takes 5,800 episodes and each episode uses 20 trajectories. The round hole takes 5,800 episodes, and each episode uses 20 trajectories. Then, the finetuned dynamics model is evaluated using a force-torque reading of 20 randomly generated trajectories.
The evaluation metric is defined as
\begin{equation}
    Err=\frac{1}{20}\sum_{t} ( \left \| \widehat{F_t}^{force}- F_t^{force} \right \|^2 +  100\left \| \widehat{F}_t^{torque}- F_t^{torque} \right \|^2 )
    \label{equ:force}
\end{equation}
where $\widehat{F_t}^{force}$ and $\widehat{F}_t^{torque}$ are the force and torque of estimation from the dynamics, and $F_t^{force}$ and $F_t^{torque}$ are the force and torque of sampled ground truth respectively.
The estimation error of the X-shaped hole is 0.094 and that of the round hole is 0.0705, which shows that our dynamics model has the ability to predict the force-torque state with high accuracy.

\subsection{Sample Efficient Transferability of the Dynamics Model}
\label{subsection1}
Next, we conduct an experiment to evaluate the sample efficiency of our dynamics model when transferring to a new peg-hole pair with unseen characteristics. A comparison is made between our pretrained dynamics model and the dynamics model learned from scratch. First, the pre-trained dynamics and an untrained dynamics model are all trained on an unseen hole in the testing set. Overall 81 points are sampled on the new hole and the percentage of data used for training varies from 2\% to 100\%. 

In Fig. \ref{fig:dynamics_result}(a)(c), the two dynamics models are trained on an ellipse hole. According to Fig. \ref{fig:dynamics_result}(a), the prediction error of the pretrained dynamics model is much lower than that from scratch using only 2$\%$ of all the data. Moreover, Fig. \ref{fig:dynamics_result}(c) shows that finetuning using approximately 20$\%$ of all the data makes the force estimation error lower than 0.1, which means the force estimation of the dynamics model is accurate. Fig. \ref{fig:dynamics_result}(b)(d) shows the result on a hard circular hole of 15 mm diameter in the test set. The pretrained dynamics model achieves much better performance than the scratch model using only 2\% of the whole 81 points. These results indicate that only a few samples collected for finetuing ensure that the pretrained dynamics can be transferred to unseen cases quickly. Our dynamics model along with grid-sampling data collection has good sample efficiency.

\subsection{Dynamics Effect on Control Policies}
To determine whether the performance of the dynamics model affects its corresponding control policies, we use the dynamics models finetuned on the 15 mm circular hole using the different quantities of data in the previous section \ref{subsection1}. Then we run MPC and model-based RL separately on these dynamics models to finish the task.
Each policy conducts 20 trials and each trial is composed of 6 steps. The Euclidean distance between the current position and the target position is recorded. Fig \ref{fig:dynamics_result}(e) shows the result of MPC and Fig \ref{fig:dynamics_result}(f) shows that of model-based RL. The x-axis is the timestep, and the y-axis is the distance. The green, red and blue lines represent the performance of policy using the dynamics models finetuned using 2\%, 20\% and 40\% of all 81 points of data, respectively. As we can see in the two figures, both MPC and model-based RL policies end with a large distance from the target hole, which means they fail to finish the task using the dynamics using 2\% data, while the dynamics using 20\% and 40\% data ensure control policies successfully insert within 3 or 4 steps. The result shows that the performance of control policies is based on the dynamics model.

\subsection{MPC on Real Robot}
To test the validity of our dynamics model on a real robot, model predictive control (MPC) is first applied to finish the peg-in-hole task. With the finetuned dynamics model for each hole in the testing set, MPC is utilized to choose the best action sequence and execute the first action on the real robot. We perform 100 inserting trials and the maximum step of each trial is limited to 6. The success rate is shown in Table. \ref{tab2}.

\subsection{Model-based Reinforcement Learning}
\label{ablation}
\vspace{-0.3\baselineskip}
Here we adopt an ablation study to show the sample efficiency of our force-torque dynamics model in the finetuning phase of model-based reinforcement learning. Two policies are initialized using our dynamics model pretrained on the training set. Afterward, the dynamics model is finetuned using 16 points sampled on the unseen hexagon hole. One policy follows our pipeline and is trained using the finetuned dynamics offline. In other words, it only interacts with dynamics rather than the real world. The other policy runs directly on the real robot and attempts to complete the hexagon peg-in-hole task. Here, we apply the same algorithm asynchronous advantage actor-critic (A3C) \cite{mnih2016asynchronous} to learn both policies. The two policies both achieve a success rate of over 90\%. We compare the number of samples needed to achieve the same success rate. The samples needed for finetuning the dynamics model on the hexagon hole are only 16, while the number of samples for finetuning the policy online is 12 k and it takes approximately 5 hours to finish the task. It is obvious that our method is more sample efficient when finetuning.

\section{CONCLUSION}
\label{sec:Conclusion}
We present a transferable force-torque dynamics model to address the peg-in-hole challenge. By training on an object dataset with different shapes, scales and elasticities, our dynamics model can be adapted quickly to new peg-and-holes with unseen characteristics. The training process is based on a novel offline data generation method that is more sample-efficient than online training on real robots. With the learned dynamics model, two control policies, model predictive control and model-based RL, can be applied to complete the insertion. Our experiments with various peg insertion tasks indicate that our dynamics model transfers well to new instances, and it can be applied to complete these tasks successfully along with different control policies. 
For future work, we would like to explore the force pattern with the introduction of the end-effector rotation. Other modalities such as visual information can also be added to enrich the observation space.

\section{ACKNOWLEDGMENT}
This work is supported in part by the National Key R$\&$D Program of China, No. 2017YFA0700800, National Natural Science Foundation of China under Grants 61772332.

\bibliography{IEEExample}

\end{document}